\documentclass[runningheads]{llncs}
\usepackage{graphicx}

\usepackage{tikz}
\usepackage{comment}
\usepackage{amsmath,amssymb} 
\usepackage{color}
\usepackage{graphicx}
\usepackage{multirow, booktabs}
\usepackage{makecell}
\usepackage{tabularx}
\usepackage{adjustbox}
\usepackage{graphicx}

\newcommand{\p}{\vspace{1mm}\noindent}
\usepackage[hidelinks]{hyperref}
\def\etal{\emph{et~al}.}
\usepackage{pdfpages}

\usepackage[accsupp]{axessibility}  

\begin{document}
\pagestyle{headings}
\mainmatter
\def\ECCVSubNumber{2663}  

\title{Hierarchically Self-Supervised Transformer for \\ Human Skeleton Representation Learning} 


\titlerunning{Hierarchically Self-Supervised Skeleton Representation Learning}
%
\author{Yuxiao Chen\inst{1} \thanks{Correspondence to: Yuxiao Chen~(yc984@cs.rutgers.edu).} 
\and
Long Zhao\inst{2} \and
Jianbo Yuan\inst{3} \and 
Yu Tian\inst{3} \and
Zhaoyang Xia\inst{1} \and
\\
Shijie Geng\inst{1} \and
Ligong Han\inst{1} \and
Dimitris N. Metaxas\inst{1} 
}
\authorrunning{Y. Chen et al.}
%
\institute{\textsuperscript{1} Rutgers University, \textsuperscript{2} Google Research, \textsuperscript{3} ByteDance Inc.
}


\maketitle

\begin{abstract}

Despite the success of fully-supervised human skeleton sequence modeling, utilizing self-supervised pre-training for skeleton sequence representation learning has been an active field because acquiring task-specific skeleton annotations at large scales is difficult. Recent studies focus on learning video-level temporal and discriminative information using contrastive learning, but overlook the hierarchical spatial-temporal nature of human skeletons. Different from such superficial supervision at the video level, we propose a self-supervised hierarchical pre-training scheme incorporated into a hierarchical Transformer-based skeleton sequence encoder (Hi-TRS), to explicitly capture spatial, short-term, and long-term temporal dependencies at frame, clip, and video levels, respectively. 
To evaluate the proposed self-supervised pre-training scheme with Hi-TRS, we conduct extensive experiments covering three skeleton-based downstream tasks including action recognition, action detection, and motion prediction. Under both supervised and semi-supervised evaluation protocols, our method achieves the state-of-the-art performance. Additionally, we demonstrate that the prior knowledge learned by our model in the pre-training stage has strong transfer capability for different downstream tasks. The source code can be found
at \textcolor{magenta}{\url{https://github.com/yuxiaochen1103/Hi-TRS}}.

\end{abstract}

\section{Introduction}
\label{sec:intro}

Human skeleton data \cite{shahroudy2016ntu,liu2019ntu,liu2017pku} are sequences of human body joints with 2D or 3D coordinates that are extracted from human activity videos. Compared with data from other modalities such as RGB frames \cite{tran2015learning,feichtenhofer2019slowfast} and depth images \cite{wang2015action,xiao2019action}, human skeletons are light-weight and more robust against variations in illumination, texture, and background~\cite{song2017end,hussein2013human}. Therefore, leveraging skeletons as the input in deep neural networks to understand human activities has become prevalent recently~\cite{vemulapalli2014human,hussein2013human,zhang2017view,song2017end,zhu2016co,li2017skeleton}.

\begin{figure}[!t]
\centering
\includegraphics[width=\linewidth]{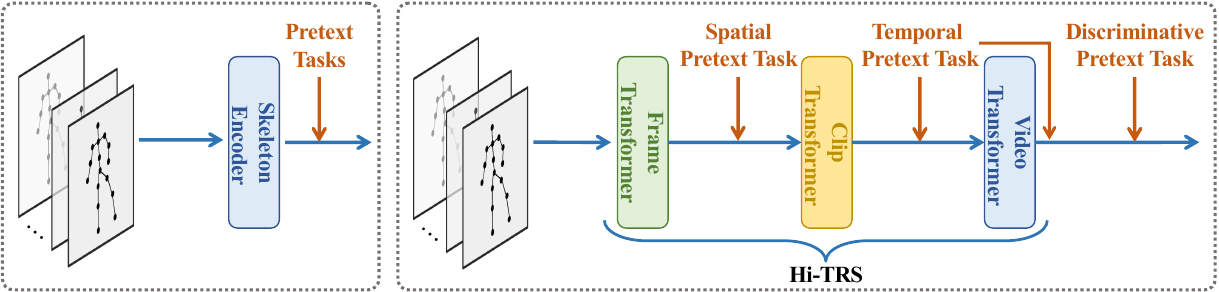}
  \caption{Comparison of pre-training strategies. \textbf{Left:} Previous methods apply pretext tasks to supervise the final output of a skeleton encoder. \textbf{Right}: We propose to hierarchically supervise outputs of the encoder at different levels during  pre-training.}
  \label{fig:method_overview}
\end{figure}

Different from other modalities, skeletons have naturally inherent spatial-temporal hierarchies. The main challenge of skeleton-based methods is how to properly capture the domain knowledge ({\em i.e.}, the correlations among the joints in the spatial and temporal domains) while extract effective feature representations from skeletons. Recent studies~\cite{yan2018spatial,shi2019two,song2017end} have achieved remarkable performance improvement by learning skeleton encoders in a fully-supervised manner. These methods require massive skeleton training data with task-specific annotations which are expensive and labor-intensive to be collected. Some studies~\cite{li20213d,su2021self,lin2020ms2l} tackle the problem by directly applying the self-supervised learning scheme designed for videos or images to skeleton data. Their pretext tasks extract video-level temporal and discriminative information but are only employed to supervise the final encoder outputs, as shown in Figure~\ref{fig:method_overview} (Left). However, these approaches do not consider the hierarchical nature of human skeletons and thus ignore the structural domain knowledge carried by them. 
 
To address the above limitations, we propose a novel skeleton representation learning framework to capture the hierarchical spatial-temporal domain knowledge of human skeletons. As shown in Figure~\ref{fig:method_overview} (Right), it consists of (1) a hierarchical Transformer-based skeleton sequence encoder, namely \emph{Hi-TRS}, incorporating with (2) a hierarchical self-supervised pre-training scheme.

Specifically, the proposed Hi-TRS models skeleton sequence in three levels. Given a skeleton sequence, the Frame Transformer \emph{(F-TRS)} and the Clip Transformer \emph{(C-TRS)} learn the spatial structures (\textbf{frame level}) and short-term fine-grained temporal dynamic dependencies (\textbf{clip level}) among the skeleton joints by applying self-attentions \cite{vaswani2017attention} on the spatial and temporal domains, respectively. Then, the clip-level embeddings are fed to the Video Transformer \emph{(V-TRS)} to summarize long-term abstract information from clips (\textbf{video level}) and produce the feature representation of the skeleton sequence. The clip-level embeddings can be applied to short-term skeleton-based tasks, such as action detection~\cite{liu2017pku,li2016online}, while embeddings from V-TRS can be used in long-term skeleton-based tasks, such as action recognition \cite{yan2018spatial} and motion prediction \cite{martinez2017human}.

Instead of only supervising the final output of the encoder as in previous work~\cite{lin2020ms2l,li20213d,su2021self}, our framework leverages different pretext tasks to supervise the encoder at different levels. As a result, the encoder acquires different types and levels of prior knowledge on human skeletons. 
To be specific, the \emph{spatial pretext task} infers the information of one joint conditioned on the other joints from the same time step. It is applied to the output of the F-TRS for learning the spatial dependencies among joints. The \emph{temporal pretext task} assists our model to capture the temporal dynamic prior by distinguishing between valid and invalid motion patterns. It supervises the outputs of C-TRS and V-TRS.
The \emph{discriminative pretext task} captures discriminative information for supervising the output of V-TRS, which enforces the model to predict future information in a contrastive manner.

We conduct extensive experiments covering a wide range of tasks and problem settings to evaluate the proposed method. Our approach outperforms the state-of-the-art skeleton representation learning methods on three downstream tasks, including \emph{action recognition}, \emph{action detection}, and \emph{motion prediction}, under both \emph{semi-supervised} and \emph{supervised learning} evaluation protocols. Most noticeably, Hi-TRS improves previous state-of-the-art methods on action recognition by 5.8\% (semi-supervised), by 8.1\% (supervised) on action detection, and by 4.2\% (4.6mm) (semi-supervised) on motion prediction. Additionally, we conclude the following key observations: (1) With the help of our hierarchical supervision, the prior knowledge learned during pre-training is more versatile to support downstream tasks at different levels than prior work using contrastive learning only on the video level (see Sections \ref{sec:exp:ad} and \ref{sec:exp:mp}); (2) Our approach demonstrates strong transfer capability under the transfer learning setting, where we achieve significant improvement on action recognition, action detection, and motion prediction tasks by 5\%, 4.5\%, and 11.7\% (12.3mm), respectively; (3) Our ablation study shows that pre-training at lower levels is beneficial to higher level downstream tasks. Interestingly, we observe similar improvement obtained on lower level downstream tasks when leveraging higher level pre-training.

\section{Related Work}
\label{sec:rel_w}

\textbf{Self-supervised Learning.} Self-supervised learning targets learning effective feature representations from unlabeled data. It trains the model to solve pre-designed pretext tasks, where labels are automatically generated from data without human efforts. Great efforts have been made in  previous work to design  pretext tasks \cite{zhang2016colorful,pathak2016context,devlin2018bert}. In computer vision, colorizing grayscale images \cite{zhang2016colorful}, image inpainting \cite{pathak2016context}, and image jigsaw puzzles \cite{noroozi2016unsupervised} are proposed to learn image feature representations. Motion prediction \cite{han2019video}, temporal jigsaw puzzle recognition \cite{noroozi2016unsupervised}, clip orders prediction \cite{xu2019self}, and sequential verification \cite{misra2016shuffle} tasks are employed to learn temporal dynamic information in videos. Recently, contrastive-based pretext tasks \cite{chen2020simple,he2020momentum} are introduced to learn instance discriminative information. On the other hand, language-based pre-training objectives are widely used in language domains~\cite{devlin2018bert,bojanowski2017enriching,pennington2014glove}. Motivated by the success of these methods, our work leverages in-domain pretext tasks to supervise the encoder at different levels.

\p \textbf{Skeleton Representation Learning.} Early skeleton representation learning methods \cite{gui2018adversarial,kundu2019unsupervised,su2020predict,zheng2018unsupervised,chenBMVC19dynamic} are mainly based on the encoder-decoder architecture. Zheng \etal{} \cite{zheng2018unsupervised} trained a GAN-based model to reconstruct the original skeleton information from the corrupted input. Su~\etal{}~\cite{su2020predict} trained the model to decode the future motion of the input skeleton sequences. Recent studies adopt the self-supervised learning schemes designed for videos or images to skeleton data. Lin~\etal{}~\cite{lin2020ms2l} trained the model to jointly solve motion prediction, temporal jigsaw puzzle, and contrastive learning discriminative tasks. Li~\etal{}~\cite{li20213d} presented a memory augmented contrastive learning framework and further improved its performance by pursuing cross-view consistency constraints. Su~\etal{}~\cite{su2021self} guided the model to learn motion consistency and continuity from videos. A shortcoming of these methods is that they do not explicitly encourage the model to learn the spatial structure of skeletons. Yang~\etal{}~\cite{yang2021skeleton} proposed to represent skeleton sequences as skeleton clouds and learn the spatial and temporal information of skeletons by solving the skeleton cloud colorization problem. However, it required training two different models to learn the spatial and temporal information, respectively. Different from these methods, we use multiple pretext tasks hierarchically to train our model so that the spatial structure, temporal dynamics, and discriminative information can be learned simultaneously.

\p \textbf{Downstream Tasks.} \emph{Action recognition} aims to predict the action category of a skeleton sequence. Studies in this area mainly focus on designing skeleton-specific architectures for feature encoding. Early methods \cite{zhang2017view,song2017end,zhu2016co,li2017skeleton,ke2017new,kim2017interpretable} applied CNNs or RNNs to extract the representation of skeleton data. Recent methods \cite{yan2018spatial,shi2019two} modeled the skeleton data as spatial-temporal graphs and extracted skeleton embeddings from graphs by Graph Convolutional Networks \cite{kipf2016semi}. More recent studies \cite{plizzari2021spatial,cheng2021hierarchical} leveraged the self-attention mechanism to extract global dependencies among joints. In this work, we use this task to evaluate the effectiveness of skeleton representation learning methods for long-term discriminative tasks.
\emph{Action detection} temporally localizes and recognizes the presence of the action in untrimmed videos~\cite{liu2017pku,song2018spatio,li2016online}. Studies in this area can be categorized into two streams. The first stream \cite{liu2017pku,li2016online} formulates the task as a frame prediction problem, and generates detection results directly from the predicted categories of each frame in a skeleton sequence. The second stream~\cite{song2018spatio,li2018co} first generates action proposals, and then recognizes action categories from them. This paper follows the first stream to evaluate skeleton representation learning methods for short-term discriminative tasks. \emph{Motion prediction} targets predicting future human poses based on a short observation of human motion \cite{zhao2019semantic,cai2021unified,barsoum2018hp,mao2021generating}. Previous methods employed RNNs to encode observed information and predict future motions~\cite{fragkiadaki2015recurrent,martinez2017human}. These models are trained to generate deterministic results. Recent work incorporated VAEs or GANs to decode multiple possible motions~\cite{barsoum2018hp,mao2021generating,cai2021unified,petrovich2021action}. To evaluate the effectiveness of learned prior knowledge, we fine-tune models to predict deterministic motion in generation tasks.

\section{Our Method}

\subsection{Hierarchical Transformer-based Encoder}

The Hi-TRS model consists of three components: F-TRS, C-TRS, and V-TRS. Given a skeleton sequence, the F-TRS first learns the spatial dependencies among the joints by applying the self-attention operation on the spatial domain. Then, the obtained results are fed to the C-TRS model to further encode the temporal fine-grained dynamics dependencies among joints and extract a feature representation at the clip level. Finally, the V-TRS infers the temporal relations among the clips and extracts the embedding of the input skeleton sequence. In the following sections, we provide details on each component.

\p \textbf{Frame Transformer (F-TRS).} Given a skeleton sequence, the positional feature of each joint is first extracted from its coordinates by a fully-connected layer with the GELU activation \cite{hendrycks2016gaussian}. F-TRS utilizes the positional features from all the joints within a frame of the skeleton sequence as input. It is composed of a stack of F-TRS layers, each of which encodes the spatial dependencies among the joints based on the self-attention mechanism.

To be specific, in the $l$-th F-TRS layer, the model starts by projecting the input feature of each joint to query, key, and value vectors \cite{vaswani2017attention} by three learnable project matrices $\mathbf{W}_{Q}^l$, $\mathbf{W}_{K}^l$, and $\mathbf{W}_{V}^l$, respectively, as described by the following equation:
\begin{equation}
	\label{eq:Q_K_V}
    \mathbf{Q}_{t}^l = \mathbf{W}_{Q}^l\mathbf{X}_{t}^{l-1}, \:
    \mathbf{K}_{t}^l = \mathbf{W}_{K}^l\mathbf{X}_{t}^{l-1}, \:
    \mathbf{V}_{t}^l = \mathbf{W}_{V}\mathbf{X}_{t}^{l-1},
\end{equation}
where $\mathbf{X}_{t}^{l-1}$ is the matrix of the input features. When $l = 1$, it consists of the positional features of the joints at the $t$-th frame; otherwise, it is the output of the previous F-TRS layer. $\mathbf{Q}_{t}^l$, $\mathbf{K}_{t}^l$ and $\mathbf{V}_{t}^l$ are the transformed outputs of query, key, and value vectors, respectively. We note that the $i$-th rows of these four matrices are correspondent to the $i$-th joint in the skeleton.

The dot-product between each pair of query and key vectors is then calculated and scaled by the dimension number of the key or value vectors. Finally, the attention weights are obtained by normalizing the scaled dot-product with a $\mathrm{Softmax}$ function. This process is defined in the following equation:
\begin{equation}
    \label{eq:att_w}
    \mathbf{A}_t^l = \mathrm{Softmax}(\frac{\mathbf{Q}_t^l {(\mathbf{K}_t^l})^T }{\sqrt{d_{k}}}), \
\end{equation}
where $(\mathbf{K}_t^l)^T$ is the transpose of $\mathbf{K}_t^l$; $d_{k}$ is the dimension number of key or value vectors; $\mathbf{A}_t^l$ is the matrix of spatial attention weights among the joints at the $t$-th frame, and its element at the $i$-th row and $j$-th column is the attention weight of the $i$-th joint with respect to the $j$-th joint.  These attention weights can be regarded as the measure of the spatial dependencies among the joints. The output feature of each joint is updated as the weighted sum of the value vectors, as shown in the following equation:
\begin{equation}
    \label{eq:weighted_sum}
        {\mathbf{X}}_{t}^l = \mathbf{A}_t^l \mathbf{V}_t^l.
\end{equation}
As a result, the spatial dependence among joints is encoded into their features. 

Following the multi-head attention mechanism \cite{vaswani2017attention}, the above  self-attention operation is performed $h$ times with $h$ different learnable projection matrices $\mathbf{W}_{Q}^l$, $\mathbf{W}_{K}^l$, $\mathbf{W}_{V}^l$, and the obtained $h$ outputs for each joint are concatenated. The results are then fed  to the Feedforward Network (FFN) \cite{vaswani2017attention}, generating the final output of the $l$-th F-TRS laye.

\p \textbf{Clip Transformer (C-TRS).} Since a skeleton sequence typically contains a large number of joints and the self-attention operation scales quadratically with respect to the number of joints, learning the fine-grained temporal dependencies over the entire skeleton sequence using self-attention is computationally expensive. To alleviate this problem, we temporally split a skeleton sequence into a sequence of clips $C$ with a sliding window. Then, the temporal dependencies among the joints within each clip of $C$ are learned using the C-TRS model.

Specifically, the input of C-TRS contains the spatial features of the joints within a clip and a [CLS] token \cite{devlin2018bert}. The [CLS] token summarizes useful information from all the joints of the clip, because its output embedding is the weighted sum of all
joints’ features, where the weights are calculated using self-attention \cite{devlin2018bert,dosovitskiy2020image}. The output of the [CLS] token from the C-TRS model is used as the feature representation of the entire clip. The C-TRS model is composed of a stack of C-TRS layers, each of which learns the temporal dependencies among joints by applying the multi-head self-attention mechanism on the temporal domain. We leverage the following equation to compute the attention weights $\mathbf{A}_c$:
\begin{equation}
    \label{eq:c_att}
    \mathbf{A}_c = \mathrm{Softmax}(\mathrm{Mask}(\frac{\mathbf{Q}_c (\mathbf{K}_c)^T}{\sqrt{d_{k}}})), \
\end{equation}
where $\mathbf{Q}_c$ and $\mathbf{K}_c$ are the matrices of the query and key vectors for all the joints within the $c$-th skeleton clip of $C$, which are generated following the same way as Equation~\ref{eq:Q_K_V}. More importantly, the $\mathrm{Mask}$ function is used for discarding spatial attention weights. It achieves this by setting the scaled dot-product among the joints from the same frame as the negative infinity, and keeps the other joints unchanged. After $\mathrm{Softmax}$, all spatial attention weights in $\mathbf{A}_c$ are equal to 0. 

The joint features are updated following the same method as in Equation \ref{eq:weighted_sum} to further encode the temporal dependencies information. The output clip-level embeddings of all the clips in $C$ are fed to the V-TRS model to extract the feature representation of the skeleton sequence.

\p \textbf{Video Transformer (V-TRS).} The V-TRS model summarizes the long-term abstracted video level information. It consists of stacked standard transformer encoder layers \cite{vaswani2017attention} and takes clip-level embeddings of all the clips in $C$ together with a [CLS] token as inputs. Each of the V-TRS layers learns the temporal dependencies among the clips. The output embedding of the [CLS] token is used as the feature representation of the skeleton sequence.

\begin{figure*}[t]
\centering
\includegraphics[width=\linewidth]{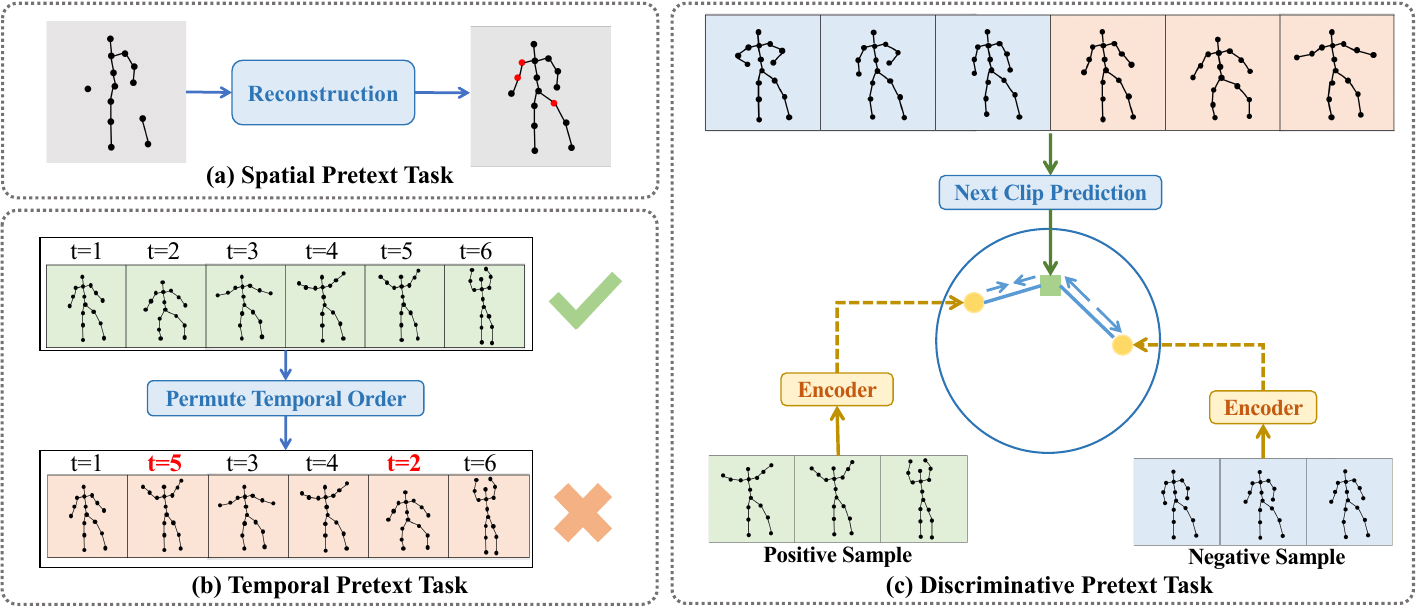}
\caption{Overview of our pre-training tasks which include: (\textbf{a}) Spatial pretext task: predicting the 3D coordinates of joints based on those of other joints from the same time step; (\textbf{b}) Temporal pretext task: predicting whether the temporal dynamic pattern of a skeleton clip or sequence is valid; (\textbf{c}) Discriminative pretext task: forecasting the embedding of the next clip of a skeleton sequence.}
\label{fig:pt_task}
\end{figure*}

\subsection{Hierarchical Self-supervised Pre-training}

In this section, we introduce the proposed pre-training tasks and describe how they can be applied to supervise the training of the proposed model.

\p \textbf{Spatial Pretext Task.~} The spatial task is to predict the coordinates of a joint based on other joints from the same time step, as shown in Figure \ref{fig:pt_task}(a). 
Given a skeleton sequence, we first randomly sample 15\% of the joints and replace the coordinate of the $i$-th sampled joints by: (1) the randomly generated coordinate 80\% of the time, (2) the coordinate randomly sampled from other joints 10\% of the time, and (3) the unchanged coordinate 10\% of the time. These strategies are inspired by the masking strategies in BERT~\cite{devlin2018bert}. The modified skeleton sequence is fed to the F-TRS model, and then the extracted spatial embeddings of the modified joints are fed to a fully-connected layer to regress their original coordinates. The model is trained to minimize the absolute error between the predicted and ground truth  coordinates by the following L1 loss $\mathcal{L}_{S}$:

\begin{equation}
    \label{eq:L_S}
    \mathcal{L}_{S} = \frac{1}{|M|} \sum_{i \in M} || \bar{y_i} - y_i ||_1,
\end{equation}
where $M$ is the set of modified joints; $|M|$ is the size of $M$;  $\bar{y_i}$ and $y_i$ are the  predicted and ground truth coordinates of the $i$-th modified joint, respectively.


\p \textbf{Temporal Pretext Task.~} The temporal task requires the model to determine whether the temporal dynamic pattern of a skeleton clip or sequence is valid, as shown in  Figure~\ref{fig:pt_task}(b). This is a binary classification problem, where the positive samples are the original skeleton sequences or the skeleton clips cropped from the original skeleton sequences, while negative samples are generated by permuting the temporal order of the positive samples. It guides the model to learn the prior knowledge of temporal dynamics.  When this task is applied to the output of C-TRS, a positive skeleton clip is generated by temporally cropping a few frames from the skeleton sequence, while the negative sample is created by swapping two randomly sampled frames of the positive clips. The output embeddings from the C-TRS model of the two clips are fed to a fully connected layer for prediction. We train the model by using the cross-entropy loss $\mathcal{L}_{T}^C$:
\begin{equation}
    \label{eq:loss_c_t}
    \mathcal{L}_{T}^C = -(log(p^+) + log(1 - p^-)),
\end{equation}
where $p^+$ and  $p^-$ are the predicted positive possibilities for the positive and negative samples, respectively.

When this task is applied to the output of V-TRS, a negative sample is generated by temporally swapping two randomly sampled clips in the skeleton clip sequence $C$. We then use a linear layer to classify whether a sample is negative or positive. The model is trained by the loss function $\mathcal{L}_{T}^V$, which follows the definition in Equation \ref{eq:loss_c_t}.

\p \textbf{Discriminative Pretext Task.~} This task predicts the embedding of the future clip of a skeleton sequence, as shown in Figure~\ref{fig:pt_task}(c). It encourages the model to learn discriminative information by supervising the task in a contrastive way. Specifically, the model is trained to predict the embedding of the last clip in $C$ based on the information from all other clips in $C$. The output from the C-TRS model of the last clip is used as the ground truth, and all other clips are fed to the V-TRS to extract a video-level embedding, which encodes the past information of the last clip. The obtained video-level embedding is fed to a fully-connected layer to regress the feature of the last clip. The model is trained by using the InfoNCE loss \cite{oord2018representation} $\mathcal{L}_{D}$:
\begin{equation}
    \label{eq:L_D}
    \mathcal{L}_{D} = \frac{exp(\bar{e_i} \cdot e_i / \tau)}{\sum_{j=1}^{B} exp(\bar{e_i} \cdot e_j / \tau)},
\end{equation}
where $\bar{e_i}$ and $e_i$ are the predicted and ground truth embedding of the last clip of the $i$-th video, respectively; $\tau$ is a temperature hyper-parameter [33]; $B$ is batch size. $\mathcal{L}_{D}$ enforces the predicted embedding of a sample to be more similar to its ground truth than to those of other negative samples. Compared with previous studies where the contrastive learning methods are based on data augmentation~\cite{lin2020ms2l,li20213d}, our method potentially requires lower computation as it does not require encoding augmented views of input data.

\p \textbf{Full Pre-training Objective}. The full objective of the proposed hierarchical self-supervised pre-training framework $\mathcal{L}_{H}$ is: $\mathcal{L}_{H} = \mathcal{L}_{S} + \mathcal{L}_{T}^{C} + \mathcal{L}_{T}^{V} + \mathcal{L}_{D}$.

\section{Experiments}

To evaluate the proposed method, we begin by introducing the datasets, evaluation protocols, and implementation details in Sections \ref{sec:dataset}, \ref{sec:protocol}, and \ref{sec:imp_detail}, respectively. We then compare our method with the state-of-the-art skeleton representation learning approaches for the action recognition, action detection, and motion prediction tasks in Sections \ref{sec:exp:ar}, \ref{sec:exp:ad}, and \ref{sec:exp:mp}, respectively. We further evaluate the transfer capability of the learned prior knowledge on human skeletons through pre-training in Section \ref{sec:exp:tranfer}. Finally, we conduct an ablation study to evaluate the proposed pre-training strategy in Section \ref{sec:alb_study}. 

\subsection{Datasets}
\label{sec:dataset}
\textbf{NTU RGB+D 60 Dataset (NTU-60).}  The NTU-60 dataset \cite{shahroudy2016ntu} contains 56,880 videos of 60 action categories. These videos are performed by 40 actors and captured by three Microsoft Kinect v2 cameras from different views. Each video contains at most two subjects. A subject has 25 joints per frame. The 3D joint locations of these joints are extracted by the  Microsoft Kinect cameras. Two common evaluation benchmarks \cite{shahroudy2016ntu} are recommended  on this dataset. In Cross-Subject (xsub) benchmark, the training videos are from 20 selected subjects, and the testing videos are from the other 20 subjects. In Cross-View (xview) benchmark,  the videos from the second and third cameras are used for training, while the videos from the first camera are used for evaluation purpose. 

\p \textbf{NTU RGB+D 120 Dataset (NTU-120).} The NTU-120 dataset \cite{liu2019ntu} is an extended version of the NTU-60 dataset. It contains 113,945 skeleton sequences from 120 action categories. There are two common protocols \cite{liu2019ntu} for this dataset. In Cross-Subject (xsub) benchmark, the samples of the selected 53 subjects are used for training, and the samples of the remaining subjects are used for testing. In Cross-Setup (xset) benchmark, the samples with even setup IDs are used for training, and those with odd setup IDs are used for testing. 

\p \textbf{PKU Multi-Modality Dataset (PKUMMD).} PKUMMD \cite{liu2017pku} is a new large-scale benchmark for continuous multi-modality 3D human action understanding. It contains almost 20,000 action instances and 5.4 million frames from 52 action categories. Actions are labeled at frame level \cite{liu2017pku}. The 3D joints are also extracted via the Microsoft Kinect v2 cameras. PKUMMD consists of two subsets: Part I and Part II. Following the common settings \cite{liu2017pku,lin2020ms2l}, the training and testing data are split under the Cross-Subject \cite{liu2017pku} protocol for each subset.

\subsection{Evaluation Protocol}
\label{sec:protocol}
Following previous work \cite{li20213d,yang2021skeleton}, our model is evaluated under two settings: (1) the supervised setting and (2) the semi-supervised setting. Under the supervised setting, the pre-trained encoder is jointly fine-tuned with a linear classifier or a LSTM-based motion decoder \cite{martinez2017human} for downstream tasks using all the labeled pre-training data. Under the semi-supervised setting, we use the same setup as the supervised setting described above except that the amount of annotated training samples used for fine-tuning is limited.

\subsection{Implementation Details}
\label{sec:imp_detail}

In the F-TRS, C-TRS, and V-TRS models, the number of their layers, attention heads, and dimensions of query vectors are all set as 2, 8, and 64, respectively. The input and output dimensions of the F-TRS, C-TRS, and V-TRS model are 128, 256, and 512, respectively. Before being fed to F-TRS, the input 3D coordinates of each joint are projected to 128 dimensions by a fully connected layer with the GELU activation \cite{hendrycks2016gaussian}.  The output of F-TRS and C-TRS are fed into a fully-connected layer to increase feature dimension to 256 and 512, respectively, before being fed into C-TRS and V-TRS.  Positional encodings \cite{vaswani2017attention} are added to the joint features or clip features to retain their spatial identity and temporal information. Specifically, standard learnable 1D positional embeddings \cite{dosovitskiy2020image} are added to the input of F-TRS and V-TRS, while learnable 2D positional embeddings \cite{dosovitskiy2020image} are used for the input of C-TRS. More details can be found in the supplementary materials.

\subsection{Results on Action Recognition}
\label{sec:exp:ar}
In this section, we evaluate our method on the action recognition task. Given a skeleton sequence, the entire Hi-TRS model is used as the encoder, and the outputs from the V-TRS model are fed into a linear classifier ({\em i.e.}, a fully-connected layer) to predict action categories. For a fair comparison with the two-streams (2S) and three-streams (3S) based methods \cite{yang2021skeleton,li20213d,su2021self}, we implement a 2S and a 3S version of our method. Specifically, we train three individual models from three different views of skeleton sequences, including joints, motions, and bones following \cite{li20213d}. During the evaluation, the 3S prediction results are obtained by fusing the prediction scores of the three models \cite{li20213d}, while the 2S prediction results are obtained by fusing the results of the joint and bone models~\cite{li20213d,su2021self}.

\begin{table}[t]

\centering

\caption{Top-1 classification accuracy (\%) for action recognition on the NTU-60 and NTU-120 datasets under the supervised setting. ``-2S'' and ``-3S'' mean two-stream and three-stream based models, respectively. The best results are highlighted in bold.}
\vspace{-10pt}
\label{tbl:super_ntu}

\resizebox{0.7\linewidth}{!}{
\begin{tabular}{lllccccc}
\toprule
\multirow{2.5}{*}{Method} &&& \multicolumn{2}{c}{NTU-60} & & \multicolumn{2}{c}{NTU-120}                         \\
\cmidrule(lr){4-5} \cmidrule(lr){7-8} 
 &&& \multicolumn{1}{c}{xsub} & \multicolumn{1}{c}{xview} & &\multicolumn{1}{c}{xsub} & \multicolumn{1}{c}{xview}\\ 
 \midrule
MS\textsuperscript{2}L \cite{lin2020ms2l} (ACMMM'20)                     && & 78.8                     & \multicolumn{1}{c}{81.8} & & - & -                         \\
VPD \cite{nie2020unsupervised} (ECCV'20)                        &&& \multicolumn{1}{c}{-}                           & \multicolumn{1}{c}{81.4}  & & - & -                         \\

MCC \cite{su2021self} (ICCV'21)             &&& \multicolumn{1}{c}{83.0} & \multicolumn{1}{c}{89.7} & & 77.0 & 77.8  \\

MCC-2S \cite{su2021self} (ICCV'21)             &&& \multicolumn{1}{c}{89.7} & \multicolumn{1}{c}{\textbf{96.3}} & & 81.3 & 83.3  \\

CrosSCLR-3S \cite{li20213d} (CVPR'21)        && & \multicolumn{1}{c}{86.2} & \multicolumn{1}{c}{92.5}  & & 80.5 & 80.4 \\
SCC-3S \cite{yang2021skeleton} (ICCV'21)           && & \multicolumn{1}{c}{88.0} & \multicolumn{1}{c}{94.9}      & & - & - \\ \midrule

Hi-TRS (Ours) &&& \multicolumn{1}{c}{86.0} & \multicolumn{1}{c}{93.0} & &80.6 &81.6    \\

Hi-TRS-2S (Ours) &&& \multicolumn{1}{c}{{89.2}} & \multicolumn{1}{c}{{95.1}} & &84.7 &86.6    \\

Hi-TRS-3S (Ours)    && & \multicolumn{1}{c}{\textbf{90.0}} & \multicolumn{1}{c}{95.7}  & &\textbf{85.3} & \textbf{87.4}                       \\
\bottomrule
\end{tabular}}
\end{table}

\p \textbf{Supervised Setting}. We compare the proposed Hi-TRS with other approaches on NTU-60 and NTU-120 under the supervised setting. The top-1 classification accuracy is reported on each benchmark. The obtained results are shown in Table \ref{tbl:super_ntu}. 
We can see that our 3S method achieves the state-of-art performance on NTU-60 and NTU-120. Note that the encoders used by several previous methods achieve better performance than our model when the parameters are randomly initialized. For example, when trained from scratch , MCC outperforms the proposed Hi-TRS by 1.9\% under the cross-subject setting on the NTU-60 dataset, and the 3S-encoders used by CrosSCLR outperforms Hi-TRS by around 3\% on the NTU-120 dataset. However, our method is able to outperform them when the models are pre-trained. These results demonstrate that the proposed hierarchical pre-training scheme enables Hi-TRS to learn powerful prior knowledge on human skeletons which can be successfully leveraged in the downstream task.

\begin{table}[t]
\centering
\caption{Top-1 classification accuracy (\%) for action recognition on the NTU-60 dataset under the semi-supervised setting. ``-2S'' and ``-3S'' mean two-stream and three-stream based models. The best results are highlighted in bold.}
\resizebox{0.8\linewidth}{!}{
\begin{tabular}{lcccccccc}
\toprule
\multirow{2.5}{*}{Method} &&& \multicolumn{2}{c}{1\% data} & \multicolumn{2}{c}{5\% data} & \multicolumn{2}{c}{10\% data} \\ 
\cmidrule(lr){4-5}\cmidrule(lr){6-7}\cmidrule(lr){8-9}
              &&& xsub         & xview         & xsub         & xview         & xsub          & xview         \\ \midrule
              
ASSL \cite{si2020adversarial} (ECCV'20)                &&& -             & -              & 57.3         & 63.6          & 64.3          & 69.8          \\

MS\textsuperscript{2}L \cite{lin2020ms2l} (ACMMM’20)          &&& 33.1         & -              & -            & -             & 65.2          & -              \\
MCC-2S \cite{su2021self} (ICCV’21)                 &&& -            & -             &  47.4         & 53.3          & 60.8          & 65.8          \\
CrosSCLR-3S \cite{li20213d} (CVPR’21)         &&& \textbf{51.1}         & 50.0          & -             &-               & 74.4          & 77.8          \\
SCC-3S \cite{yang2021skeleton} (ICCV’21)               &&& 48.3         & \textbf{52.5}          & 65.7         & 70.3          & 71.7          & 78.9          \\ \midrule
Hi-TRS (Ours)                 &&& 39.1         & 42.9          & 63.3         & 68.3          & 70.7          & 74.8          \\

Hi-TRS-3S (Ours)                &&& 49.3         & 51.5          & \textbf{71.5}         & \textbf{74.8}          & \textbf{77.7}          & \textbf{81.1}          \\ \bottomrule
\end{tabular}}
\label{tbl:semi_ntu}
\vspace{-10pt}
\end{table}
\p \textbf{Semi-supervised Setting}. Following the standard setup in \cite{lin2020ms2l,li20213d,yang2021skeleton}, we fine-tune our pre-trained encoder and the randomly initialized linear classifier with randomly sampled 1\%, 5\%, and 10\% of the training data on the NTU-60 dataset, respectively. From the results reported in Table \ref{tbl:semi_ntu}, we observe that the proposed Hi-TRS outperforms the state of the art by a large margin under the 5\% and 10\% settings. On the other hand, we note that our model performs slightly worse under 1\% setting. We hypothesize this is due to the fact that 1\% of the training data is insufficient to train Transformer-based encoders with a large number of parameters as explained in \cite{vaswani2017attention}.

\begin{figure*}[t]
\centering
\includegraphics[width=\linewidth]{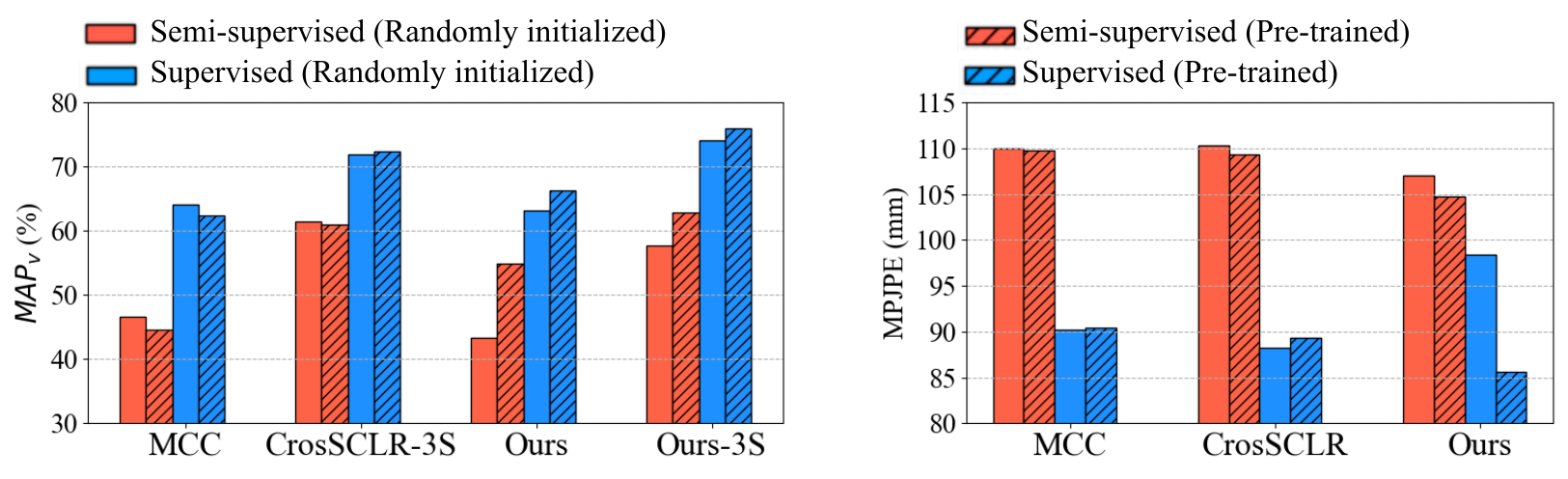}
\caption{\textbf{Left}: mAP\textsubscript{v} (\%) results on the action detection task (the higher the better). \textbf{Right}: MPJPE (mm) results on the motion prediction task (the lower the better). We note that the reported results of both MCC~\cite{su2021self} and CrosSCLR~\cite{li20213d} are based on our implementation. Please refer to the supplementary material for the implementation details and exact numbers of each model.}
\label{fig:ad_and_md}
\end{figure*}

\subsection{Results on Action Detection}

\label{sec:exp:ad}

In this section, we compare our method with previous approaches on the action detection task. This experiment aims to evaluate the effectiveness of the learned skeleton representations for short-term discriminative tasks.

We formulate the action detection task as a per-frame classification problem following the setting in  \cite{liu2017pku,li2016online}. Given one certain frame, we extract a short clip that contains its surrounding information from the entire skeleton sequence. (Due to space limitation, please refer to the supplementary material for more details on how video clips are extracted). The obtained video clip is then fed into F-TRS and C-TRS to extract its feature representation. Finally, a linear classifier is applied to predict the action category of the input frame based on the obtained feature representation.

Following the evaluation setting of~\cite{li20213d,li2016online,shi2019two}, the experiments are conducted on PKUMMD Part I subset. According to~\cite{liu2017pku}, we adopt mAP\textsubscript{v} (mean average precision of different videos) and mAP\textsubscript{a} (mean average precision of different actions) with the overlapping ratio of 0.5 as the evaluation metrics. The experimental results of the mAP\textsubscript{v} metric are presented in Figure \ref{fig:ad_and_md} (Left). From this figure, we can find that our method outperforms previous approaches under both supervised and semi-supervised settings. More importantly, we find that MCC underperforms its randomly initialized encoder by 2.1\% and 1.8\% in the supervised and semi-supervised settings, respectively. Meanwhile, CrosSCLR-3S also underperforms its randomly initialized encoder by 0.5\% in the semi-supervised setting. One possible reason is that these two methods focus on learning long-term temporal representations \cite{su2021self,li20213d}.
As a result, their learned prior knowledge is not effective for short-term downstream tasks. In contrast, the proposed Hi-TRS surpasses its randomly initialized counterpart by a large margin. This demonstrates that our method can capture powerful prior knowledge for short-term downstream tasks, thanks to the proposed hierarchical pre-training strategy. We also have the same observations when the mAP\textsubscript{a} metric is utilized, and please refer to the supplementary material for the corresponding results and qualitative analysis.

\subsection{Results on Motion Prediction}
\label{sec:exp:mp}
In this task, the model is trained to predict the motions in the future 400 milliseconds based on an observation of two seconds, following the short-term motion prediction protocol defined in \cite{martinez2017human}. We adopt this task to evaluate the effectiveness of learned prior knowledge for generation tasks.

Specifically, the observed skeletons are fed into the proposed Hi-TRS to extract feature representations. The outputs of the V-TRS model are fed into a GRU-based decoder \cite{martinez2017human} to predict the joint coordinates of skeletons for the future 400 milliseconds. The model is then trained to minimize the Euclidean distance between the predicted poses and ground truths. 

Following previous work \cite{cai2021unified,zhao2019semantic}, we employ MPJPE (mm) as the evaluation metric, which measures the distance between the ground truths and the generated results. The experiments are conducted on the NTU-60 cross-subject benchmark as shown in Figure \ref{fig:ad_and_md} (Right). From this figure, we can find that our method outperforms previous methods by a large margin under both supervised and semi-supervised settings. Additionally, the learned prior knowledge of the previous methods is not useful under the semi-supervised setting. On the other hand, our method significantly outperforms the randomly initialized counterpart under different settings. It is consistent with the observations on the action detection task, demonstrating that our learned prior knowledge is more versatile to support different downstream tasks than the previous approaches. We also provide qualitative results in the supplementary material.

\begin{table}[t]
\centering
\caption{Results of motion prediction, action recognition, and action detection under the transfer learning setting. The best results are highlighted in bold.} 
\resizebox{0.8\linewidth}{!}{
\begin{tabular}{lcccccccc}
\toprule
\multirow{2.5}{*}{Pre-training Dataset}  && \multicolumn{2}{c}{PKU Part II}   && \multicolumn{2}{c}{PKU Part I} \\
\cmidrule(l){3-4} \cmidrule(l){6-7} 
 &&MPJPE $\downarrow$ & Accuracy $\uparrow$ && mAP\textsubscript{a} $\uparrow$  & mAP\textsubscript{v} $\uparrow$     \\ \midrule
Randomly Initialized Encoder                                                                            && 105.4 & 50.9     && 53.4         & 63.2          \\ 
NTU-60-xsub                                                                        &&94.2 & 55.0     && 55.2         & 66.6        \\
NTU-120-xsub                                                        &&\textbf{93.1} & \textbf{55.9}     && \textbf{57.9}         & \textbf{67.3}        \\ \bottomrule
\end{tabular}}
\label{tbl:transfer}
\vspace{-10pt}
\end{table}

\subsection{Evaluation of Transfer Learning}
\label{sec:exp:tranfer}
In this section, we evaluate whether the learned knowledge of Hi-TRS through the pre-training process is transferable across datasets. To this end, we first pre-train two encoders under the cross-subject protocol on NTU-60 and NTU-120, respectively. The pre-trained encoders are then fine-tuned on PKUMMD Part I and PKUMMD Part II for action detection, action recognition, and motion prediction. The obtained results are then compared with the ones of a randomly initialized encoder. These results are reported in Table \ref{tbl:transfer}. We can observe that pre-training can improve performance for different-level downstream tasks by a large margin, because the learned prior knowledge is transferable and versatile. Additionally, from the results of ``NTU-120-xsub'' and ``NTU-60-xsub'', we find that pre-training on larger datasets can further improve transfer capability.

\subsection{Ablation Study}
\label{sec:alb_study}

\begin{table}[t]
\centering
\caption{Results of the ablation study under the supervised setting on the NTU-60 cross-subject benchmark for action recognition and motion prediction. ``-'' means the encoder's parameters  are randomly initialized. F, C, and V mean that the pre-training tasks are applied on the output of F-TRS, C-TRS, and V-TRS, respectively. The best results are highlighted in bold.}
\resizebox{0.76\linewidth}{!}{
\begin{tabular}{@{}lccccccccccccccccc@{}}
\toprule
Pre-trained Level & & & - & & F    && C   & & V   & & F+C &  & F+V  & & C+V  & &  F+C+V         \\ \midrule
Accuracy(\%) $\uparrow$   & & &  79.6 & & 80.8 && 81.1 && 82.0  & & 83.9 & & 84.1 & & 84.0 &  & \textbf{86.0} \\ 
MPJPE(mm) $\downarrow$       &  &  & 98.4 &  & 97.3 && 96.7 && 88.1  &  & 95.4 & & 87.4    &  & 90.2  &  & \textbf{85.6} \\ \bottomrule
\end{tabular}}
\label{tbl:albation_ntu}
\vspace{-10pt}
\end{table}

\begin{table}[t]
\centering
\caption{Results of the ablation study under the supervised setting on the PKUMMD Part I subset for action detection. ``-'' means that the encoder's parameters are randomly initialized.  F, C, and V mean that the pre-training tasks are applied on the output of F-TRS, C-TRS, and V-TRS, respectively. The best results are in boldface.}
\resizebox{0.44\linewidth}{!}{
\begin{tabular}{lccccccc}
\toprule
Pre-trained Level &&&  -  &&   F+C &&  F+C+V  \\ 
\midrule
mAP\textsubscript{a} &&& 53.4 && 55.6 && \textbf{58.4} \\
mAP\textsubscript{v} &&& 63.2 && 65.1 && \textbf{66.3} \\
\bottomrule
\end{tabular}}
\label{tbl:albation_pku}
\end{table}

In this section, we evaluate the effectiveness of the proposed hierarchical pre-training strategy. This is achieved by comparing the performance of the encoders that are pre-trained on different levels.


We first show how pre-training on low levels affects the performance of the high-level downstream tasks. The experiments are conducted on the NTU-60 cross-subject benchmark for action recognition and motion prediction under the supervised protocol. The obtained results are reported in Table \ref{tbl:albation_ntu}. We find that pre-training on each level (frame level, clip level, and video level) can achieve performance improvement over the randomly initialized encoder, thanks to the powerful prior knowledge learned from the pre-training tasks of each level. Additionally, pre-training on any combination of two levels achieves higher performance improvement than pre-training on only one level. 
More importantly, the best improvement is achieved when the encoder is pre-trained on all levels. This confirms the fact that our full model manages to combine prior knowledge containing spatial structure, temporal dynamics, and discriminative information for human skeletons during the pre-training stage. 

To further explore how the high-level pre-training tasks affect the low-level downstream tasks, we conduct experiments on the PKUMMD Part I subset for action detection. The results are shown in Table \ref{tbl:albation_pku}. Please refer to the supplementary material for the results of more model variants.  We can see that pre-training on high level (video level) leads to performance improvement on the low level downstream task as well, since it can introduce temporal dynamic information and complementary discriminative information.  


\section{Conclusion}
In this work, we proposed a novel method that encodes skeleton sequences using a hierarchical Transformer-based encoder and designed a pre-training scheme consisting of three pretext tasks at three different levels. We conducted extensive experiments under different learning settings. For the  supervised and semi-supervised settings, our method achieves the state-of-the-art performance against competitive baselines. Moreover, the learned prior knowledge through hierarchical pre-training shows strong transfer learning capability for downstream tasks at different levels. The experimental results demonstrate that our method is an effective way for learning feature representations of skeleton data.

\section{Acknowledgments}

This work has been funded partly by NSF IUCRC CARTA-1747778, 2235405, 2212301, 1951890, and 2003874 to Dimitris N. Metaxas.




%
%
\bibliographystyle{splncs04}
\bibliography{egbib}



\section*{Supplementary material}

# Hierarchically Self-Supervised Transformer for Human Skeleton Representation Learning (Supplementary Material)

Yuxiao Chen[1], Long Zhao[2], Jianbo Yuan[3], Yu Tian[3], Zhaoyang Xia[1], Shijie Geng[1], Ligong Han[1], and Dimitris N. Metaxas[1]

[1] Rutgers University, [2] Google Research, [3] ByteDance Inc.

## 1 Implementation Details of Pre-training

The implementation details for pre-training are introduced in this section.

**Data Augmentation.** During training, data augmentations are applied to both the spatial and temporal domains of a skeleton sequence. Specifically, the spatial augmentation method includes: (1) shearing [3]: the skeleton of each frame is sheared at a random angle by applying a linear transformation matrix whose elements are uniformly sampled from the range [-0.5, 0.5]; (2) adding noise: a noise sampled from the uniform distribution with the range [-0.2, 0.2] is added to 5 randomly sampled joints at each frame. The temporal augmentation method contains: (1) time interpolation [6]: it generates a new skeleton sequence by linearly interpolating the positions of each joint between consecutive time with a random scale whose range is from 0 to 1; (2) temporal cropping: it temporally crops a random portion of frames from the original sequence as the training sample. Note that for each training sample, only one spatial and one temporal augmentation method are used.

**Model Pre-training.** We pre-train the model using the Nvidia A100 GPU. The batch size is set to 256. The dropout rate [7] is set to 0.2, and the L2 weight decay is set to 0.01. The Adam optimizer [2] with the learning rate of $4 \times 10^{-4}$ is used for training. We use linear learning rate warmup for the first 1,000 iterations followed by polynomial decay. The model is trained for 30,000 iterations. The window size, stride, and dilate of the sliding window are set to 7, 7, and 2, respectively. For the training samples which contain two subjects, we divide them into two samples, each of which contains the skeleton sequence of one subject.

## 2 Implementation Details of Downstream Tasks

This section describes the implementation details for each downstream task.

**Action Recognition.** For the action recognition task, we use the entire Hi-TRS model as the encoder. The outputs from the V-TRS model are fed to a linear classifier (a fully-connected layer) to predict action categories. To train

our network, the cross-entropy loss and the Adam Optimizer with a learning rate of $4 \times 10^{-4}$ are used. The batch size is set to 128. The window size, stride, and dilate of the sliding window are set as 7, 7, and 2, respectively. If the encoder is pre-trained, the learning rate is divided by 2 at the 20th, 25th, and 30th epochs. The training process is terminated at the 50th epoch. If the encoder is randomly initialized, the learning rate is divided by 2 at the 40th, 50th, and 60th epochs. The training process is terminated at the 70th epoch.

**Action Detection.** In the action detection task, each skeleton sequence contains multiple different actions [4]. It is formulated as a per-frame classification problem following the setup in [4]. Specifically, if a frame belongs to an action of the $c$-th category, its label is $c$; otherwise, its label is 0. For a skeleton sequence with $n$ frames, a sliding window with size 7, stride 1, and dilate 4 is employed to temporally crop $n$ clips from the sequence. The $n$-th obtained clip contains about one-second surrounding information of the $n$-th frame. It is fed to the F-TRS and C-TRS models to extract its clip-level embedding. Then, a linear classifier (a fully-connected layer) is used for predicting the action category of the $n$-th frame. The models are trained to minimize the cross-entropy loss by using the Adam Optimizer. The learning rate and batch size are set as $4 \times 10^{-4}$ and 256, respectively. The model is trained for 5 epochs.

**Motion Prediction.** The model used in this task consists of an encoder and a decoder. The encoder is our proposed Hi-TRS. For the decoder, we use the LSTM-based architecture proposed in [5], and we implement it based on its Pytorch implementation [1]. We train the model to minimize the mean squared error loss. We set the batch size as 128 and set the dropout rate of the decoder as 0.5. We train the model for 25 epochs by using the Adam Optimizer. The initial learning rate is set to $1 \times 10^{-4}$ and is decayed by 0.5 at the 20th epoch.

## 3  Implementation Details of Previous Methods

Previous work [8,3] only evaluated their methods on the action recognition task. The reported results for the action detection and motion prediction tasks are based on our implementation. In this section, we introduce the implementation details of these methods.

**MCC [8].** Since the official implementation of this method is not publicly available, we implement it by ourselves according to the details in [8]. Specifically, the ST-GCN encoder [9] and momentum memory bank [1] used by MCC are implemented based on their officially released codes [2] [3]. During pre-training, we completely follow the hyper-parameters settings of MCC [8], such as the batch size, memory bank size, and learning rate. When conducting downstream tasks, the pre-trained encoders are fine-tuned following the same settings as our method for a fair comparison.

---

[1] https://github.com/enriccorona/human-motion-prediction-pytorch
[2] https://github.com/yysijie/st-gcn
[3] https://github.com/facebookresearch/moco

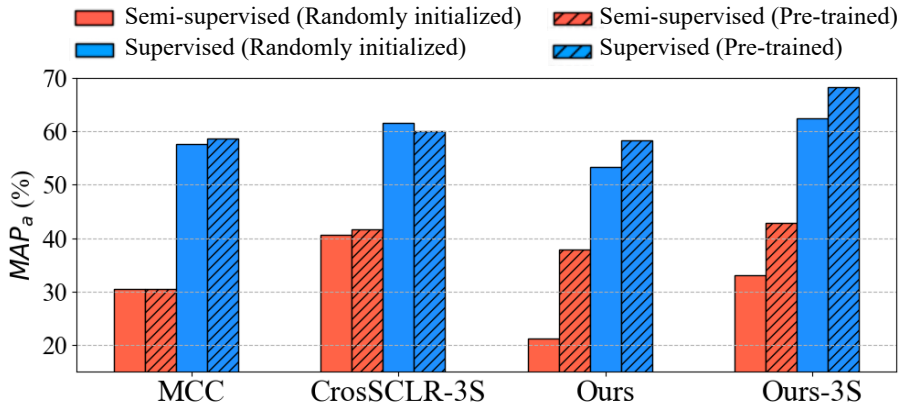

**Fig. 1.** $mAP_a$ (%) results on the action detection task.

**Table 1.** Results (%) for the action detection task on the PKUMMD Part I subset under supervised and semi-supervised settings.

| Method | 100% data | | 10% data | |
|---|---|---|---|---|
| | $mAP_a$ | $mAP_v$ | $mAP_a$ | $mAP_v$ |
| MCC | 58.6 | 62.3 | 30.5 | 44.5 |
| CrosSCLR-3S | 60.1 | 72.4 | 41.6 | 60.9 |
| Ours | 58.4 | 66.3 | 37.9 | 54.8 |
| Ours-3s | **68.2** | **76.0** | **42.9** | **62.9** |

**CrosSCLR [3].** We pre-train the encoders by using its official implementation[4], and fine-tune the pre-trained encoders following the same settings as ours for action detection and motion prediction.

## 4  Additional Results of Action Detection

The results for action detection when using the $mAP_a$ metric are shown in Figure 1. We can observe the similar findings when the $mAP_v$ metric is used (shown in the main paper). Additionally, the exact numbers are reported in Table 1.

**Qualitative analysis.** We show two action detection examples in Figure 2. We can see that the temporal boundaries of the actions detected by our methods are more accurate than the baselines (MCC and CrosSCLR-3S). The results further demonstrate that the prior knowledge learned through our different-level pretext

---

[4] https://github.com/LinguoLi/CrosSCLR

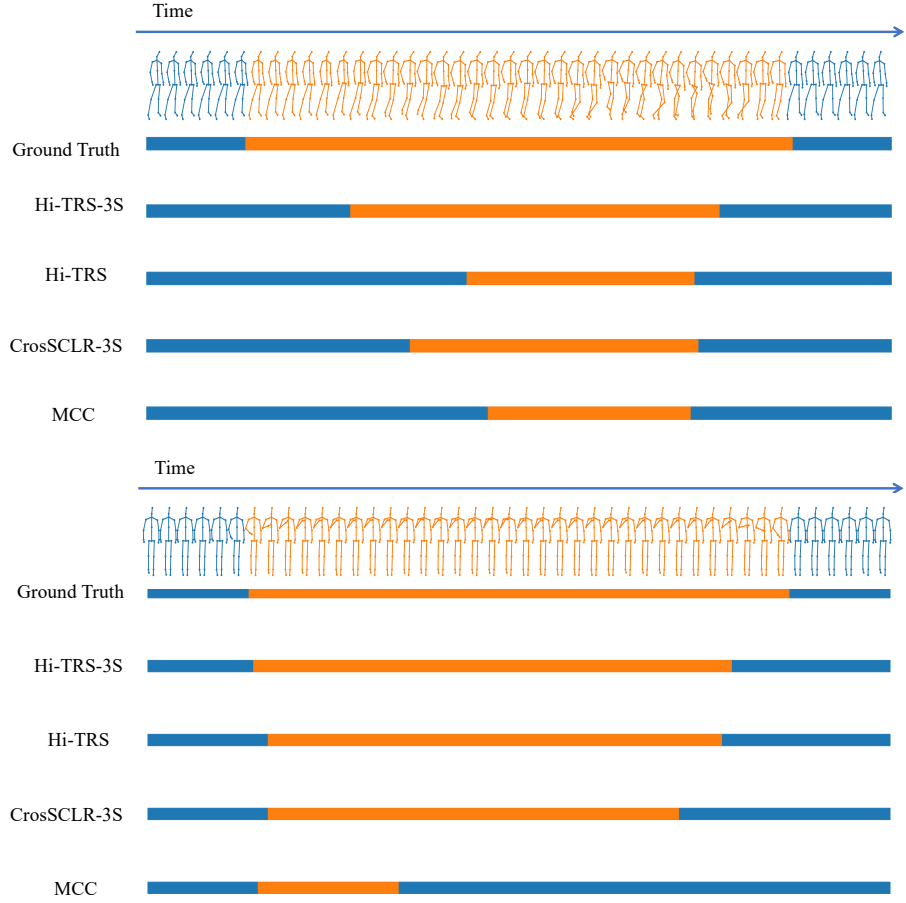

**Fig. 2.** Examples for the action detection task on the PKUMMD Part I subset. The blue regions represent that there are no actions. The orange regions denotes that the correspondent actions are detected. **Top:** An example showing the action detection results of the one foot jumping action. **Bottom:** An example showing the action detection results of the touching chest action.

tasks is more useful for extracting short-term information than that learned by using the baselines' video-level pretext tasks.

## 5   Additional Results of Motion Prediction

The correspondent numbers of the main paper's Figure 3 Right are reported in Table 2.

**Qualitative analysis.** We visualize two cases of motion prediction in Figure 3. We can observe that the MCC and CrosSCLR models fail to predict the

**Table 2.** Results for the motion prediction task on the NTU-60 cross-subject benchmark under the supervised and semi-supervised settings.

| Method | 100% data MPJPE ↓ | 10% data MPJPE ↓ |
|---|---|---|
| MCC | 90.4 | 109.8 |
| CrosSCLR | 89.3 | 109.3 |
| Ours | **85.6** | **104.7** |

**Table 3.** Results (%) of the ablation study for action detection

| Pre-trained Level | - | F | C | V | F+C | F+V | C+V | F+C+V |
|---|---|---|---|---|---|---|---|---|
| mAP$_a$ | 53.4 | 55.5 | 55.3 | 55.9 | 55.6 | 57.9 | 56.3 | **58.4** |
| mAP$_v$ | 63.2 | 64.1 | 63.9 | 63.9 | 65.1 | 64.0 | 64.0 | **66.3** |

fine-grained movement patterns, such as the motion of the legs and right arm in Figure 3 **Top** and **Bottom**, respectively. This may because they are only pre-trained to learn coarse video-level motion patterns [8] or discriminative information [3]. By contrast, our proposed hierarchical pre-training framework help our encoders capture the motion patterns at different level. As a result, our method can produce more accurate motion prediction.

## 6   Additional Results of Ablation Study

We report the results of all model variances for action detection in Table 3. We observe that pre-training at lower level is beneficial to higher level downstream tasks. It is consistent with the ablation study for other downstream tasks (see Table 4 of the paper). Additionally, the results show that leveraging higher level pre-training also improves lower level downstream tasks.

## References

1. He, K., Fan, H., Wu, Y., Xie, S., Girshick, R.: Momentum contrast for unsupervised visual representation learning. In: Proceedings of the IEEE/CVF Conference on Computer Vision and Pattern Recognition. pp. 9729–9738 (2020)
2. Kingma, D.P., Ba, J.: Adam: A method for stochastic optimization. arXiv preprint arXiv:1412.6980 (2014)
3. Li, L., Wang, M., Ni, B., Wang, H., Yang, J., Zhang, W.: 3d human action representation learning via cross-view consistency pursuit. In: Proceedings of the IEEE/CVF Conference on Computer Vision and Pattern Recognition. pp. 4741–4750 (2021)
4. Liu, C., Hu, Y., Li, Y., Song, S., Liu, J.: Pku-mmd: A large scale benchmark for continuous multi-modal human action understanding. arXiv preprint arXiv:1703.07475 (2017)

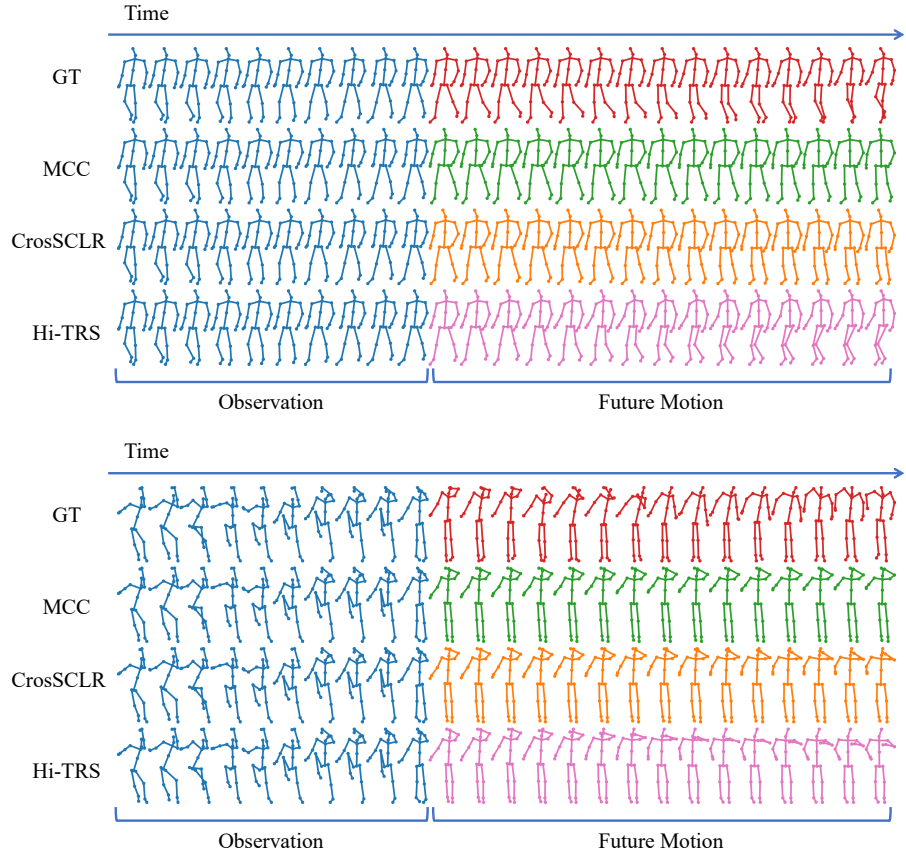

**Fig. 3.** Examples for the action prediction task on the NTU-60 cross-subject benchmark under the supervised setting. "GT" means ground truth.


\end{document}